\documentclass[letterpaper,10pt,conference]{ieeeconf}

\IEEEoverridecommandlockouts
\usepackage[nosort]{cite}
\usepackage{amsmath,amssymb,amsfonts}
\usepackage{graphicx}
\usepackage{booktabs}
\usepackage{multirow}
\usepackage{xcolor}
\usepackage{url}
\usepackage[hidelinks]{hyperref}
\usepackage{cleveref}

\newcommand{\colhead}[1]{\multicolumn{1}{c}{\begin{tabular}[b]{@{}c@{}}#1\end{tabular}}}
\newcommand{\colheadlast}[1]{\multicolumn{1}{c@{}}{\begin{tabular}[b]{@{}c@{}}#1\end{tabular}}}

\crefname{figure}{Fig.}{Figs.}
\Crefname{figure}{Fig.}{Figs.}
\crefname{section}{Sec.}{Secs.}
\Crefname{section}{Sec.}{Secs.}
\crefname{table}{Table}{Tables}
\Crefname{table}{Table}{Tables}

\title{Multi-Link Safety Filtering for VLA Policies Around Moving Hazards}

\author{Yatharth Agarwal\\School of Electrical and Computer Engineering\\Purdue University, West Lafayette, IN, USA\\{\tt\small agarw414@purdue.edu}
\and
Vijay Raghunathan\\School of Electrical and Computer Engineering\\Purdue University, West Lafayette, IN, USA\\{\tt\small vr@purdue.edu}%
\thanks{This work was supported in part by CoCoSys and CogniSense, two of the seven centers in JUMP 2.0, a Semiconductor Research Corporation (SRC) program sponsored by DARPA.}}

\AddToHook{para/begin}[squeeze]{\looseness=-1}

\begin{document}

\maketitle
\thispagestyle{empty}
\pagestyle{empty}

\begin{abstract}
A vision--language--action (VLA) policy can finish a manipulation task while knocking over objects unrelated to it, so task success alone does not show that the policy is safe to deploy in clutter. We study how to keep a pretrained VLA policy clear of such hazards at run time without retraining it, which requires guarding more of the arm than the end effector, following the hazard as it moves, and sharing onboard compute with the policy. Our training-free shield covers the gripper, wrist, and forearm with five ellipsoids and filters every commanded motion through one barrier program against a keep-out ellipsoid fitted from RGB-D perception at reset. Sparse optical flow then carries that ellipsoid's center along with the hazard, with no repeated detection or refitting. Over six simulated hazard-motion conditions, the shield lowers collision from $65.62\%$ to $27.27\%$ and raises safe-success, task completion without collision, from $29.35\%$ to $50.43\%$. Ablations show that guarding the arm links protects beyond end-effector shielding, and that tracking recovers most of the protection lost when the hazard estimate is frozen at reset. On heterogeneous edge hardware, the five-ellipsoid barrier runs on the CPU in $2.2$~ms at the 99th percentile, and trimming the vision--language prefix and taking fewer flow-matching steps shortens each $\pi_{0.5}$ policy call on the integrated GPU from $343$ to $177.3$~ms. On a physical SO-101 arm across four tasks, the arm touched the hazard in 3 of 16 shielded episodes versus 11 of 16 unshielded ones. Project page: \url{https://yathag.github.io/multilink-safety-filter/}
\end{abstract}

\section{Introduction}\label{sec:intro}

Robot manipulation is shifting from task-specific controllers to general policies that follow language instructions: vision--language--action (VLA) policies act across diverse tasks and scenes~\cite{pi05,pi0,openvla}, and multimodal large language models now command robot arms directly from camera images~\cite{rt2}. In shared workspaces, task completion alone is insufficient: a robot reaching for a dish may complete the task while its forearm knocks over a nearby glass. The commanded goal does not account for the cost of disturbing that unrelated object, and near sharp tools or people, the same contact can cause injury. With moving hazards, the $\pi_{0.5}$~\cite{pi05} flow-matching policy completes $62.93\%$ of episodes but disturbs the hazard in $65.62\%$ of episodes. Protecting against such a hazard requires constraints that extend beyond the task objective, and those constraints must follow it as it moves.

\suppressfloats[t]
\begin{figure}[t]
  \centering
  \includegraphics[width=\linewidth]{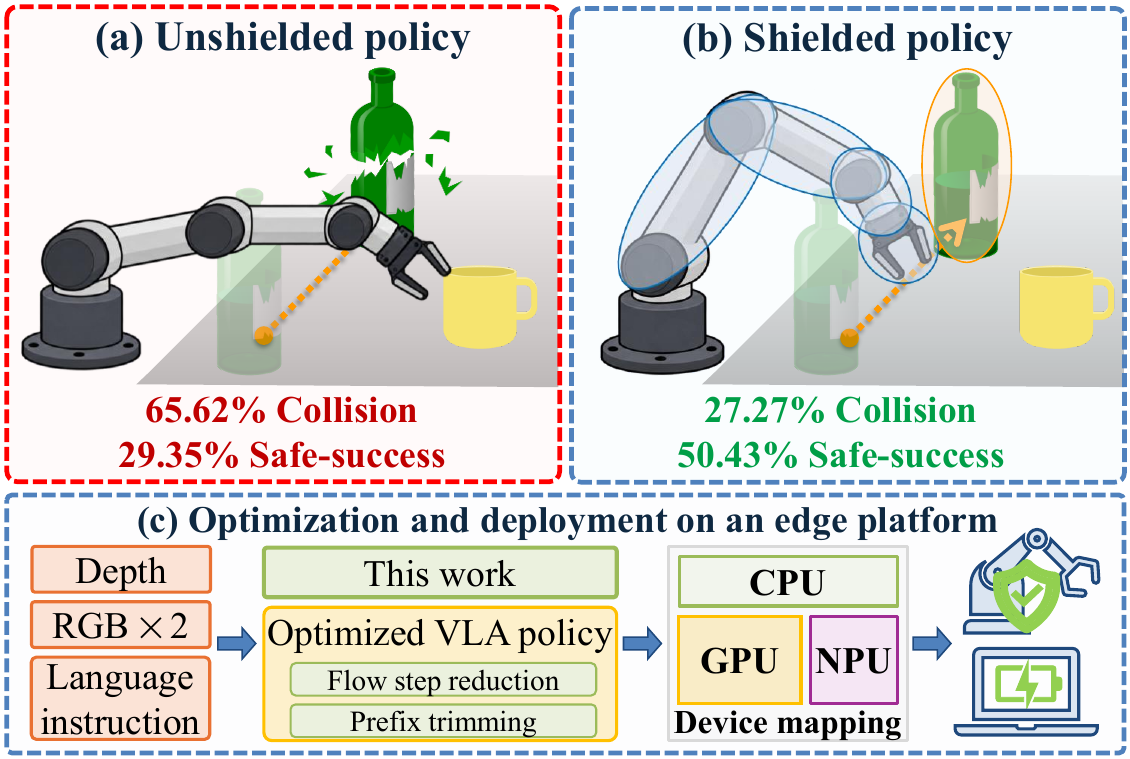}

  \caption{\textbf{Multi-link shielding around a pretrained policy.} (a,b) The unshielded policy and our shield on the moving-hazard benchmark in simulation. (c) Policy optimization and a measured device mapping place the complete loop on one heterogeneous edge platform.}
  \label{fig:overview}
  \vspace{-2em}
\end{figure}

Fine-tuning on collision-free demonstrations can encode avoidance in a policy's weights~\cite{libero_safety}, but it depends on demonstrations that represent the hazard. A runtime safety filter instead imposes a keep-out constraint on a hazard designated at execution time, without retraining; control-barrier filters modify the nominal command to satisfy it~\cite{ames_cbf,cbf_qp}, and existing VLA safety filters establish this interface~\cite{vlsa_aegis,agsf_knows}. Because the filter acts only on commanded motion, it is compatible with diverse safety-fine-tuned policies or a multimodal language model.

Deployment, however, requires protection across the arm links that sweep the workspace, updates to the keep-out region as it moves, and execution within a compute budget shared with the policy. Onboard execution keeps control and its protection independent of network availability and latency. These requirements are coupled: broader coverage protects only if the hazard estimate is current, and keeping it current adds computation to the control loop.

We present a runtime shield that addresses these requirements alongside a pretrained policy, without changing its weights. At reset, a vision-language model (VLM) names the hazard, an open-vocabulary detector localizes it, and registered depth yields a three-dimensional ellipsoid. The barrier program guards five ellipsoids across the gripper, wrist, and forearm, and sparse optical flow transports the hazard ellipsoid's center as the hazard moves, without re-detection or refitting. On a laptop processor with a CPU, an integrated GPU (iGPU), and a neural processing unit (NPU), the policy runs on the iGPU, the fastest device measured for it, while the millisecond-scale barrier and tracker stay on the CPU.

The evaluation compares the complete system against the unshielded policy and runtime-filter baselines on the static-hazard SafeLIBERO benchmark~\cite{vlsa_aegis} and a new moving-hazard benchmark. Collision captures protection, task success captures task performance, and safe-success captures both. Ablations isolate the effects of arm coverage, hazard estimation, tracking, and policy configuration, while edge measurements characterize the cost of running the shield alongside the policy.

These requirements pose a single question: can a fixed VLA policy be protected from a task-irrelevant moving object beyond its end effector, with a hazard estimate that stays current, within an onboard compute budget? The contributions answer it along three axes: coverage, motion, and deployment:

\begin{enumerate}
  \item \textbf{Coverage: multi-link runtime shielding for pretrained policies.} Five guarded ellipsoids extend the end-effector barrier across the gripper, wrist, and forearm by adding one constraint per element to a single barrier program. Broader coverage reduces collisions under image-based perception without a significant task-success cost.
  \item \textbf{Motion: hazard transport without re-detection, and a moving-hazard benchmark.} Sparse optical flow transports the hazard ellipsoid without refitting. A benchmark of six hazard-motion conditions tests protection: our shield reduces collisions from $65.62\%$ to $27.27\%$ and increases safe-success from $29.35\%$ to $50.43\%$.
  \item \textbf{Deployment: heterogeneous edge characterization and mapping.} The policy call is measured on the CPU, the iGPU, and the NPU, and the complete loop then runs under the latency-minimizing mapping: policy, detector, and hazard namer on the iGPU, barrier and tracker on the CPU. Every policy configuration is evaluated with the shield active, and the same mapping drives a physical SO-101 arm around a tracked moving hazard.
\end{enumerate}

\section{Related Work}\label{sec:related}

\textbf{Safety filters for learned manipulation.}
Control barrier functions enforce forward invariance under their model assumptions by constraining a nominal controller through a quadratic program~\cite{ames_cbf,cbf_qp}; robust variants incorporate bounded state-estimation error~\cite{mrcbf}. Recent systems impose related constraints on VLA policies by fitting a semantic barrier at reset~\cite{vlsa_aegis}, grounding constraints from policy attention~\cite{agsf_knows}, or enforcing them within the action decoder~\cite{nsym_flow,barrier_flow}. However, end-effector constraints leave the wrist and forearm unguarded as they move through the scene. Manipulator barrier filters guard the links with sphere models~\cite{oscbf,sampled_wholebody_cbf}, and Any-Body Guard addresses robot geometry in configuration space but uses a quasi-static scene representation~\cite{anybody_guard}. We instead guard five ellipsoids spanning the gripper, wrist, and forearm against one motion-updated hazard ellipsoid.

\textbf{Perception and motion.}
A geometric safety filter is only as current as its hazard estimate. Open-vocabulary detection with registered depth can construct that estimate from images~\cite{groundingdino}, but motion can invalidate a reset-time fit. Per-step segmentation can relocate the centers of hazard ellipsoids fitted at reset~\cite{agsf_knows}, and promptable video segmentation maintains object identity through streaming inference~\cite{sam2}; both run a neural network at every update, sharing the edge platform's compute budget with the policy. Sparse optical flow~\cite{lucaskanade} instead transports the fitted ellipsoid, preserving its reset shape.

\textbf{Safety training and runtime constraints.}
Fine-tuning on collision-free demonstrations encodes avoidance behavior in the policy weights and adds no online filtering cost~\cite{libero_safety}. Runtime shielding complements this learned behavior by accepting a newly designated hazard at execution time and imposing an explicit geometric constraint without updating the policy. The scope here is task-irrelevant hazards that should remain untouched; tasks requiring contact with the hazard call for constraints beyond a strict keep-out region.

\textbf{Edge deployment.}
Efficient VLA systems reduce backbone size, input tokens, or action-decoding cost~\cite{smolvla,edgevla,openvla_oft}. For $\pi_{0.5}$, which reuses an encoded vision--language prefix during iterative action generation~\cite{pi05,pi0}, work on iterative-head policies, adaptive multi-view pruning, quantization, and cross-accelerator characterization is particularly relevant~\cite{bfapp,quantvla,xpu_char}. Policy efficiency alone does not determine the performance of a shielded system: hazard estimation and filtering impose workloads at different cadences, and policy approximations alter the commands presented to the shield. The deployment question is therefore how to allocate shared compute while retaining protection and task performance. This work maps perception, tracking, filtering, and policy inference to heterogeneous edge hardware at their respective cadences, and the evaluation tests each policy configuration with runtime shielding active.

\section{System Design}\label{sec:pipeline}

\begin{figure*}[t]
  \centering
  \includegraphics[width=\textwidth]{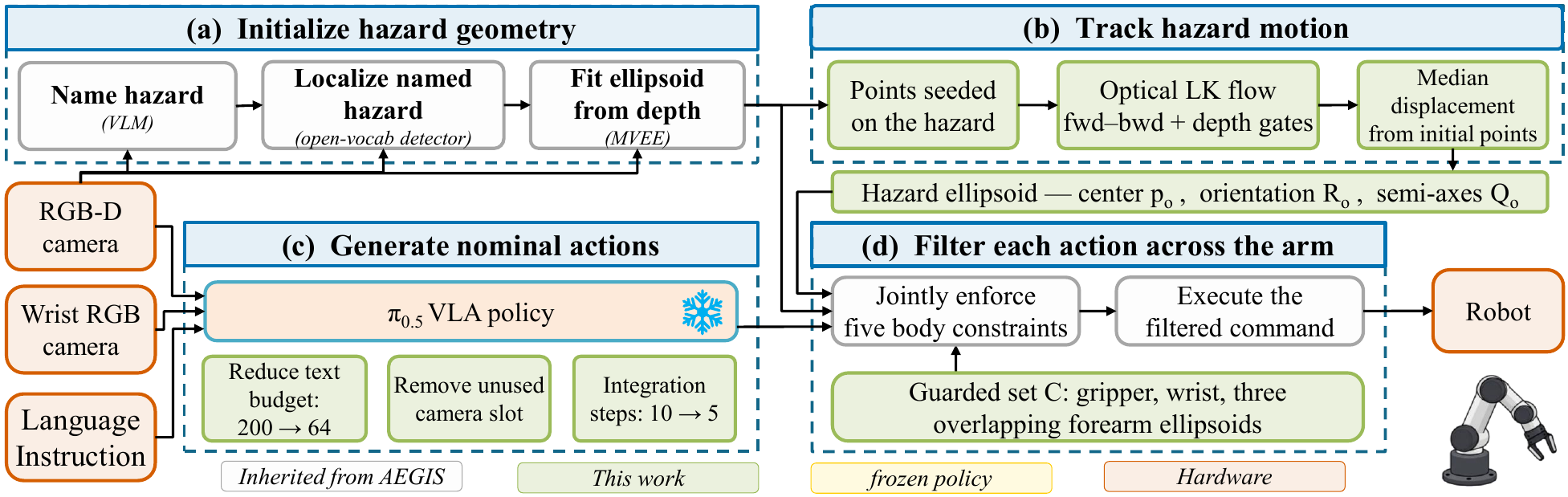}

  \vspace{-0.5em}

  \caption{\textbf{The shield runs perception and control at distinct cadences.} (a) At reset, perception names the hazard, detects it, and fits its ellipsoid. (b) Sparse Lucas--Kanade (LK) optical flow transports the ellipsoid center every five steps. (c) The policy returns 10 actions per call; the panel shows the optimized configuration. (d) Five guarded ellipsoids constrain each command at nominal $20$ Hz. Perception and control exchange only the hazard ellipsoid.}
  \label{fig:pipeline}
  \vspace{-1.5em}
\end{figure*}

Our shield is a training-free runtime filter between a pretrained VLA policy and the robot. The shield acts only on the policy's commanded end-effector motion, so it does not depend on the policy's internal architecture. The evaluation uses the $\pi_{0.5}$ flow-matching policy~\cite{pi05,pi0}, whose structure \cref{ssec:edgelevers} exploits for edge execution.
Building on AEGIS~\cite{vlsa_aegis}, we retain that system's semantic hazard naming, open-vocabulary grounding, depth-derived ellipsoid, and ellipsoid-separation control-barrier formulation. This common interface carries three extensions: protection across the arm's distal links, transport of the fitted barrier when the hazard moves, and execution of the complete loop on heterogeneous edge hardware. The policy returns a chunk of normalized end-effector commands; each command is filtered against the measured robot state immediately before execution. Perception and control exchange only the hazard ellipsoid, allowing motion updates without changing the barrier formulation (\cref{fig:pipeline}).

\subsection{The Barrier Program}
\label{ssec:qp}

Let $\mathcal{C}$ index the guarded ellipsoids, each of which contributes a keep-out constraint. The barrier program, a control-barrier-function quadratic program (CBF-QP), jointly adjusts the commanded motion to satisfy these constraints while minimizing the deviation of the commanded motion from the policy's nominal command. The hazard is represented by one ellipsoid with center $p_{\mathrm{o}} \in \mathbb{R}^{3}$, rotation $R_{\mathrm{o}} \in SO(3)$, and semi-axes $Q_{\mathrm{o}} = \operatorname{diag}(q_{1},q_{2},q_{3})$ with $q_i>0$,

\vspace{-1.2em}
 
\begin{equation}
  \mathcal{O} \;=\; \bigl\{\, y \in \mathbb{R}^{3} \;:\;
  \lVert Q_{\mathrm{o}}^{-1} R_{\mathrm{o}}^{\top}
  ( y - p_{\mathrm{o}} ) \rVert_{2} \le 1 \,\bigr\}.
  \label{eq:keepout}
\end{equation}

For each $k\in\mathcal{C}$, the guarded ellipsoid $\mathcal{B}_k(x)$ of the form \eqref{eq:keepout}, with center $p_k(x)$, rotation $R_k(x)$, and semi-axes $Q_k$, is rigidly attached to its link; the constraint protects this modeled volume. Following rotating-hyperplane barriers~\cite{rotating_hyperplane_cbf}, an auxiliary unit vector $z_k$ selects a plane tangent to $\mathcal{B}_k(x)$, and the barrier $h_k(x,z_k)$ is the minimum signed distance from that plane to $\mathcal{O}$, positive on the side away from $\mathcal{B}_k(x)$; $h_k\geq0$ certifies that the plane separates the two ellipsoids. The map $G_k(x)=J_k(x)J_{\mathrm{ee}}^{+}(x)$, with $J_k$ and $J_{\mathrm{ee}}$ the Jacobians at the center of $\mathcal{B}_k$ and at the end effector and $(\cdot)^{+}$ the pseudoinverse, relates the commanded end-effector twist $u\in\mathbb{R}^{6}$ to motion at the center of $\mathcal{B}_k$, assuming the minimum-norm joint velocity $\dot{q}=J_{\mathrm{ee}}^{+}u$ realizes $u$. The simulated arm instead executes $u$ through an operational-space controller, whose null-space motion admits a different $\dot{q}$, so $G_k$ predicts rather than reproduces the link motion. At each control step, the inherited program solves

\vspace{-1.2em}

\begin{equation}
  \begin{aligned}
    \min_{u,\,\dot{z}}\;\; & \lVert u-u_{\mathrm{nom}}\rVert^{2}_{W}
      + \lVert \dot{z}-\dot{z}_{\mathrm{nom}}\rVert^{2}_{W_{z}} \\
    \text{s.t.}\;\; & a^{u}_{k} G_{k}(x)\,u + a^{z}_{k}\dot{z}_{k}
      + \alpha\, h_{k}(x,z_{k}) \ge 0, \;\; \forall k \in \mathcal{C}.
  \end{aligned}
  \label{eq:qp}
\end{equation}

The controller executes the minimizing $u^{\star}$ and integrates $\dot{z}^{\star}$ over the control step, renormalizing each $z_k$ to unit length. Here $x$ is the robot configuration, $u_{\mathrm{nom}}$ is the policy's nominal end-effector twist, $z$ and $\dot{z}$ stack the vectors $z_k$ and their rates, and $\dot{z}_{\mathrm{nom}}$ stacks the reference rates inherited from AEGIS.

The coefficients $a^u_k\in\mathbb{R}^{1\times6}$ and $a^z_k\in\mathbb{R}^{1\times3}$ are the partial derivatives of $h_k$ with respect to the pose of $\mathcal{B}_k$ and to $z_k$, evaluated at the measured state. The positive-definite weighting matrices $W$ and $W_z$ penalize deviations from the nominal inputs, and $\alpha=10$ sets the linear barrier gain.

\subsection{Multi-Link Coverage}
\label{ssec:coverage}

An end-effector barrier leaves the swept volume of proximal links outside its modeled safe set. The design therefore instantiates $\mathcal{C}$ with five guarded ellipsoids distributed over the gripper, wrist, and forearm. The elongated forearm link is represented by three overlapping ellipsoids that retain the link's full cross-section and span its long axis; the remaining ellipsoids are derived directly from the links' collision geometry. The elbow and upper arm carry no guarded ellipsoid.

Through $G_k$, each ellipsoid's row in \eqref{eq:qp} constrains its own motion rather than treating it as rigidly attached to the gripper. All five rows share the same hazard ellipsoid in a single barrier program, which adds one constraint and one auxiliary vector per ellipsoid while keeping the objective, solver, and hazard representation.

\subsection{Hazard Initialization and Tracking}
\label{ssec:tracking}

\textbf{Reset-time initialization}
The VLM names the hazard from the task instruction and external image, using a short list of candidate hazard phrases when applicable. An open-vocabulary detector grounds the phrase in each camera view, and the selected boxes crop the registered depth images. Workspace, outlier, and density filtering precede a minimum-volume enclosing ellipsoid (MVEE) fit. The selected detection fixes the object identity for the episode. 

\textbf{Motion updates}
For a static hazard, the reset ellipsoid is held fixed. For a moving hazard, the shield retains its reset orientation and semi-axes, and only its center is transported. Let $p_{\mathrm{o}}^{0}$ be the fitted center, $c(t)$ a three-dimensional tracker position, and $c^{0}$ its value at the reset measurement. The update is
$p_{\mathrm{o}}(t)=p_{\mathrm{o}}^{0}+(c(t)-c^{0})$, with $R_{\mathrm{o}}(t)=R_{\mathrm{o}}^{0}$ and $Q_{\mathrm{o}}(t)=Q_{\mathrm{o}}^{0}$. Using an anchored displacement rather than $c(t)$ as an absolute center estimate cancels a fixed projection offset, while retaining one hazard shape avoids frame-to-frame geometric variation. Center transport therefore assumes that the hazard translates without large rotation or change in its visible shape.

The tracker operates in the fixed external camera. At initialization, the fitted ellipsoid's center is projected onto the image, and a regular grid of nearby pixels is retained only where the measured depth agrees with the center depth. These persistent points are propagated by Lucas--Kanade flow~\cite{lucaskanade}. Forward--backward agreement and depth consistency retain points that continue to support the hazard, and the median displacement from each point's seed position yields a robust image-plane translation without summing incremental center updates. The camera ray passing through the translated image center intersects the horizontal plane through $p_{\mathrm{o}}^{0}$ to obtain $c(t)$, assuming the hazard maintains its reset height. Two gates reject an unsupported translation: a bound on the center displacement per measurement, and a threshold on the normalized correlation with the reset appearance. When either gate fires, the center is held at its last accepted position while a bounded template search looks for the hazard; a match re-seeds the tracker. 

The tracker measures this displacement every $s=5$ control steps and holds the center between measurements, so the hazard ellipsoid updates within an action chunk.

% \textbf{Guarantee scope}
% Assuming a fixed hazard, exact state and geometry, an initially safe state, and feasible continuous-time execution under the modeled dynamics, the barrier conditions preserve the modeled safe set $\mathcal{S}=\{(x,z):h_k(x,z_k)\geq0,\ \forall k\in\mathcal{C}\}$. That result holds for the program \eqref{eq:qp}, not the deployed loop. The loop estimates geometry from images and depth, holds each command over a control step, predicts link motion using the approximate map $G_k$, and periodically transports the hazard ellipsoid. Each step solves the static program at the hazard's latest estimated pose, so the loop produces a sequence of static certificates rather than one certificate for the moving case. Between updates, the constraint carries no hazard-velocity term, so the hazard can close a margin the program believes it holds. At an update, the ellipsoid moves while the arm does not, so $h_k$ jumps, and the value certified at the end of one interval is not the value that opens the next. When the program is infeasible, the implementation runs the unfiltered action and logs it, because stopping is not certified against a hazard that continues to approach (\cref{sec:conclusion}). Collision reduction is therefore an empirical result, and no invariance guarantee is claimed for the deployed loop. A robust formulation could tighten constraints with a perception-error bound~\cite{mrcbf}, account for sample-and-hold error~\cite{sampled_wholebody_cbf}, and carry a hazard velocity in a time-varying barrier.

\textbf{Guarantee scope}
Under the standard assumptions of a fixed hazard, exact state and geometry, an initially safe state, and feasible continuous-time execution under the modeled dynamics, the barrier conditions in \eqref{eq:qp} preserve the modeled safe set
$\mathcal{S}=\{(x,z):h_k(x,z_k)\geq0,\ \forall k\in\mathcal{C}\}$.
The deployed system relaxes these assumptions: geometry is estimated from RGB-D observations, commands are applied in sample-and-hold fashion, link motion is approximated through $G_k$, and the hazard pose is updated periodically from optical flow. Consequently, each control step certifies the barrier conditions with respect to the latest estimated hazard pose rather than providing a continuous-time invariance guarantee for the moving-hazard loop. In particular, motion between updates and discrete changes in the estimated hazard pose are not represented explicitly in \eqref{eq:qp}. If the program becomes infeasible, the current implementation executes and logs the nominal action rather than invoking a certified fallback. We therefore treat collision reduction in the deployed system as an empirical result. Extending the shield with perception-error bounds, sampled-data barrier conditions, and an explicit hazard-velocity term would provide a path toward stronger guarantees under moving hazards.

\subsection{Edge-Aware Policy Execution}
\label{ssec:edgelevers}

Each policy invocation encodes the current vision--language prefix once~\cite{pi05,pi0}, then repeatedly integrates the action expert's flow-matching field. Attention is bidirectional within the prefix, but prefix tokens do not attend to action tokens, so the prefix keys and values are computed once per call and reused across integration steps. A new observation changes the prefix, so the cache does not carry over across calls. Each integration step attends to the cached prefix, so redundant prefix slots increase both the one-time encoding work and the repeated prefix-action attention, whereas reducing the number of integration steps acts directly on the repeated path.

Two changes trim the prefix, and a third shortens the integration loop, none of them touching the policy weights. First, the benchmark provides two camera views, while the serving configuration allocates a third, masked camera slot. Masking does not skip that slot's image encoder or its $256$ prefix positions, so the deployed configuration removes it before encoding. Second, the padded text budget falls from $200$ to $64$ tokens; the longest benchmark instruction uses $21$ tokens. This prefix trimming reduces the prefix from $3\times256+200=968$ to $2\times256+64=576$ positions. Third, the integration steps fall from ten to five, and the resulting approximation is evaluated with the shield active.

Workload size and cadence decide the mapping. The barrier program and the optical-flow update are millisecond-scale, issued at every nominal $50$ ms control step and every fifth step, respectively. Each is shorter than the overhead of an accelerator dispatch, so both run on the CPU. Dense policy inference, issued once per ten-action chunk, requires a dedicated accelerator such as an iGPU or an NPU. 

\section{Benchmarks and Metrics}\label{sec:protocol}

\begin{table*}[t]
  \centering
  
  \caption{\textbf{Static and moving-hazard performance.} C/S/SS: collision/task success/safe-success (\%); bold: best per row in (a); underline: best per block in (b). EE: end-effector; both coverage variants hold the hazard estimate frozen at reset. Shield-inactive episodes stay in every denominator. Distances: mm; Mean: over the six conditions. $^{\S}$Adapted from the authors' code (\cref{sec:protocol}).}
  \label{tab:moving}
  \label{tab:baseline_conditions}
  \label{tab:coverage}
  \vspace{-0.5em}
  
  \footnotesize
  \setlength{\tabcolsep}{3pt}
  \begin{minipage}[t]{0.69\textwidth}
    \centering
    \textbf{(a) Moving benchmark: C$\downarrow$ / S$\uparrow$ / SS$\uparrow$}\par\smallskip
    \begin{tabular}{@{}lrrrr@{}}
      \toprule
      Condition & \colhead{$\pi_{0.5}$ (unshielded)} & \colhead{AEGIS$^{\S}$} & \colhead{OSCBF$^{\S}$} & \colheadlast{Ours} \\
      \midrule
      Escape 300 & 45.99 / 67.63 / 42.15 & 29.17 / 62.66 / 46.63 & 31.73 / \textbf{69.87} / 53.53 & \textbf{22.44} / 64.58 / \textbf{56.41} \\
      Escape 150 & 53.04 / \textbf{65.38} / 39.58 & 40.22 / 53.69 / 37.50 & 41.83 / 64.90 / 45.67 & \textbf{24.52} / 58.49 / \textbf{49.04} \\
      Escape 50 & 72.76 / 56.41 / 23.88 & 42.47 / 58.17 / 39.42 & 57.69 / \textbf{58.65} / 33.49 & \textbf{25.32} / 55.29 / \textbf{44.07} \\
      Shuttle 300 & 59.62 / 68.59 / 37.18 & 41.67 / 63.14 / 43.91 & 44.71 / \textbf{68.75} / 47.92 & \textbf{36.54} / 62.82 / \textbf{51.92} \\
      Orbit 25 & 78.85 / 66.03 / 18.59 & 38.14 / \textbf{74.68} / \textbf{50.32} & 59.94 / 68.43 / 36.06 & \textbf{33.65} / 66.83 / 48.72 \\
      Stationary & 83.49 / 53.53 / 14.74 & 22.76 / \textbf{62.98} / 51.60 & 57.85 / 60.42 / 34.62 & \textbf{21.15} / 60.26 / \textbf{52.40} \\
      \midrule
      Mean & 65.62 / 62.93 / 29.35 & 35.74 / 62.55 / 44.90 & 48.96 / \textbf{65.17} / 41.88 & \textbf{27.27} / 61.38 / \textbf{50.43} \\
      \bottomrule
    \end{tabular}
  \end{minipage}\hfill
  \begin{minipage}[t]{0.29\textwidth}
    \centering
    \textbf{(b) Static SafeLIBERO}\par\smallskip
    \begin{tabular}{@{}lrrr@{}}
      \toprule
      Method & \colhead{C$\downarrow$} & \colhead{S$\uparrow$} & \colheadlast{SS$\uparrow$} \\
      \midrule
      $\pi_{0.5}$ (unshielded) & 83.44 & 60.00 & 15.06 \\
      AEGIS$^{\S}$ & 30.75 & \underline{64.88} & 49.75 \\
      OSCBF$^{\S}$ & 66.88 & 60.56 & 27.88 \\
      Ours & \underline{27.00} & 60.31 & \underline{51.69} \\
      \midrule
      Multi-link, frozen & \underline{25.19} & 61.62 & \underline{52.88} \\
      EE coverage & 31.31 & \underline{65.62} & 51.56 \\
      \bottomrule
    \end{tabular}
  \end{minipage}
  \vspace{-2em}
\end{table*}

The evaluation covers static and moving hazards in two simulated benchmarks that share the simulator, robot, policy interface, and evaluation metrics. Image-based configurations estimate hazard geometry from rendered RGB and registered simulator depth; they do not use the exact hazard pose or shape. 

\textbf{Static hazards.} SafeLIBERO~\cite{vlsa_aegis} extends the LIBERO suites~\cite{libero} with a designated hazard object and a collision criterion for each scene. The evaluation comprises $32$ scenes: four suites $\times$ two hazard levels $\times$ four tasks. The static benchmark runs every scene, with $1{,}600$ episodes per configuration. It compares our shield with the unshielded policy and with two runtime safety methods, AEGIS and OSCBF. Two coverage variants, end-effector-only and multi-link, share one perception pipeline and controller.

\textbf{Moving hazards.} A new moving-hazard benchmark reuses these scenes and metrics, with a commanded hazard trajectory defined over the episode. Each moving configuration runs $624$ episodes per condition, with $48$ initial states across $13$ scenes. The orbit uses a separate set of $13$ scenes on which a $25$ mm orbit is feasible; nine of them are shared with the other conditions. The six conditions are a stationary hazard, a circular orbit of $25$ mm radius, a recurrent $300$ mm shuttle, and one-way escape trajectories of $50$, $150$, and $300$ mm that hold at their endpoints. All moving trajectories use a commanded speed of $50$ mm/s. Each episode records the hazard's true pose, commanded pose, and per-camera visibility at every step.

Reported mean rates average the six hazard-motion conditions. The repository baselines run all six conditions with the same episode keys (\cref{sec:results}). The ablations (\cref{sec:ablations}) and the policy-configuration comparison (\cref{sec:edge}) use the $300$ mm escape condition, whose reset-to-final displacement is the largest, so a stale hazard estimate loses the most over the episode. Every simulated configuration uses the public $\pi_{0.5}$ LIBERO checkpoint (\texttt{pi05\_libero})~\cite{pi05}. It runs in its released serving configuration with three camera slots, $200$ text tokens, and ten integration steps. Edge timing and energy use the reduced configuration of \cref{ssec:edgelevers}, whose effect on outcomes \cref{sec:edge} is measured with the shield active.

\textbf{Metrics.} Safety is measured through collision rate and reported alongside task success and safe-success, defined as task completion without collision. For comparability with prior work, the evaluation adopts the SafeLIBERO displacement-based collision criterion of the released AEGIS benchmark code~\cite{vlsa_aegis}, written \emph{collision} throughout: an episode is flagged when the sum of absolute hazard position differences along the three axes exceeds $1$ mm at any control step in MuJoCo. Displacement is measured from the initial hazard position in static scenes and from the latest commanded position in moving scenes, excluding scripted motion from the latter. The criterion records a disturbance of the hazard, which need not involve robot--hazard contact, so every episode also records MuJoCo robot--hazard contacts. Across the six hazard-motion conditions, $82.05\%$ of the unshielded policy's flagged episodes carry a recorded contact, against $48.29\%$ under our shield: shielding removes contact episodes before it removes light disturbances. Rates on this stricter endpoint are therefore reported alongside the displacement endpoint (\cref{sec:performance}).

For the static benchmark, a replay of the recorded actions of the unshielded, coverage-variant, and complete-system configurations recovers peak hazard displacement and recomputes collision at $10$ mm to assess sensitivity to collision severity. 

\textbf{Implementation}
The simulated robot is the Franka Panda of the LIBERO suites in MuJoCo. The hazard namer is Qwen3-VL, and the detector is Grounding DINO~\cite{groundingdino}, prompted with the benchmark's original caption, an oracle-corrected caption, or the VLM-generated name used in deployment. A fixed scale factor converts the policy's normalized delta-pose actions into the end-effector twists the program uses, and converts the filtered twist back before execution. The barrier program \eqref{eq:qp} is solved with OSQP through CVXPY, weighting the twist deviation by $W=I_{6}/25$ and each guarded ellipsoid's auxiliary rate by $W_z=I_{3}$.

\textbf{Baseline implementations}
AEGIS~\cite{vlsa_aegis} and OSCBF~\cite{oscbf} run from the authors' public code, integrated with the same policy, simulator, and benchmark episodes. AEGIS retains its end-effector barrier and two-view geometry fitted at reset, with its GLM-4.5V hazard namer replaced by Qwen3-VL. OSCBF retains torque-level filtering and its robot-sphere model, configured to match the simulated robot's dynamics and torque limits, and takes hazard geometry and motion from our original caption perception pipeline. The robot envelopes therefore differ across methods. OSCBF keeps its published spheres and gains without tuning; its spheres end short of the open fingertips, whereas the AEGIS hand ellipsoid and ours enclose the gripper with margin.

\textbf{Statistical comparisons.} All configurations run the same episodes with the same policy sampling seed, so comparisons are paired. Episodes from a single scene are correlated, so per-episode differences are averaged within each scene ($13$ moving, $32$ static) and these scene averages serve as the samples; a mean over conditions exchanges the two configurations of a scene jointly in every condition that contains it, across the $17$ scenes the six conditions span. A difference is significant when a paired permutation test across the scene averages yields $p<0.05$ after Holm--Bonferroni correction~\cite{holm} for the differences claimed in the same sentence. Two configurations are equivalent when the $90\%$ confidence interval of their difference lies within $\pm3$ percentage points, a margin set before the experiments.

\textbf{Timing and energy.} The control loop targets $20$ Hz. After warmup, filter latency is measured per control step and policy latency per chunk invocation; the median and the $99$th percentile (p50 and p99) are reported at these respective cadences. Policy-configuration comparisons run on the development platform, where call times are medians of within-episode percentiles; edge timing is measured on a Dell XPS 14 with an Intel Core Ultra X7 358H (Panther Lake), an iGPU, and an NPU. Energy and average power are read from the system-scope RAPL \texttt{psys} counter, a platform total with no per-device attribution, and are reported per isolated invocation above the $14.5$ W idle power.

% \footnote{OpenVINO \texttt{2025.4.1}; PyTorch \texttt{2.9.1+xpu}; NPU driver-compiler \texttt{1.32.1.20260422}; CVXPY \texttt{1.9.2} with OSQP \texttt{1.1.3}.} 

All reported rates are computed in simulation. The physical SO-101 evaluation in \cref{sec:edge} reports episode counts instead, with an external Intel RealSense D455 feeding both the policy and the shield. Each episode is scored by observation: a success when the object reaches its target, and a contact when the arm touches the hazard.

\section{Evaluation}\label{sec:results}

This section first compares safety and task performance across static and moving hazards, then evaluates edge execution and policy optimizations with the shield active. AEGIS and OSCBF are external reference points that keep their own robot envelopes and gains, so this comparison establishes practical standing.

\begin{table}[t]
  \centering
  \caption{\textbf{Execution on the edge platform.}
  (a) In-loop medians per policy call and its prefix encoding, cache, and integration loop on the CPU, iGPU, and NPU.
  (b) Shield components.
  Power: mean above the idle floor over an isolated invocation; Energy: above idle, per isolated invocation.}
  \label{tab:edge}
  \vspace{-0.5em}

  \footnotesize
  \setlength{\tabcolsep}{2pt}
  \textbf{(a) Policy call by device}\\[0.4ex]
  \begin{tabular*}{\linewidth}{@{\extracolsep{\fill}}l*{6}{r}@{}}
    \toprule
    Device & \colhead{Call\\(ms)} & \colhead{Encode\\(ms)} & \colhead{Cache\\(ms)} & \colhead{Loop\\(ms)} & \colhead{Power\\(W)} & \colheadlast{Energy\\(J)} \\
    \midrule
    CPU  & 7844.2          & -              & -              & -              & 38.5          & 296 \\
    iGPU & \textbf{177.3}  & \textbf{33.2}  & \textbf{78.6}  & \textbf{65.2}  & 44.0          & \textbf{7.86} \\
    NPU  & 446.2           & 135.0          & 242.5          & 68.7           & 23.8          & 8.84 \\
    \bottomrule
  \end{tabular*}

  \vspace{0.6ex}

  \textbf{(b) Shield components on the deployed mapping}\\[0.4ex]
  \begin{tabular*}{\linewidth}{@{\extracolsep{\fill}}lllrr@{}}
    \toprule
    Component & \colhead{Device} & \colhead{Cadence} & \colhead{p50 / p99 (ms)} & \colheadlast{Energy (J)} \\
    \midrule
    Barrier (5 elements) & CPU  & step     & 2.0 / 2.2       & 0.0189 \\
    Optical flow         & CPU  & stride 5 & 3.1 / 3.6       & 0.132  \\
    Detector (2 views)   & iGPU & reset    & 457.8 / 549.4   & 19.7   \\
    Hazard namer         & iGPU & reset    & 1146.0 / 1153.8 & 51.1   \\
    \bottomrule
  \end{tabular*}
  \vspace{-1.5em}
\end{table}

\subsection{Safety and Task Performance}
\label{sec:performance}

Our shield reduces collisions and increases the rate of safe task completion around moving hazards. Collision falls from $65.62\%$ to $27.27\%$ and safe-success rises from $29.35\%$ to $50.43\%$ (\cref{tab:moving}a); task success, $62.93\%$ unshielded and $61.38\%$ with the shield, shows neither a significant change nor equivalence within the $\pm3$ percentage-point margin. Against AEGIS, our shield collides significantly less often, in $27.27\%$ of episodes compared to $35.74\%$, while its safe-success rate of $50.43\%$, compared to $44.90\%$, is not significantly higher.

OSCBF also reduces collisions relative to the unshielded policy, but exceeds our shield's collision rate by $8.2$ to $36.7$ percentage points in all six conditions. With the original-caption perception, OSCBF collides in $48.96\%$ of episodes and reaches $41.88\%$ safe-success, both significantly worse than ours; its task success of $65.17\%$, compared to $61.38\%$, shows neither a significant difference nor equivalence. The OSCBF gap does not show that torque-level filtering protects less than action-level filtering, because the robot envelopes differ. With the hazard enclosed in a bounding sphere instead of the fitted ellipsoid, OSCBF collides in $25.80\%$ of $300$ mm escape episodes, and none of its three rates differ significantly from those of the complete system.

Counting only episodes with a recorded robot--hazard contact lowers every rate and leaves the ranking of the four systems unchanged: across the six conditions, collision is $53.85\%$ unshielded, $31.86\%$ for OSCBF, $23.61\%$ for AEGIS, and $13.17\%$ for our shield.

On static SafeLIBERO, our shield collides in $27.00\%$ of episodes, compared to $30.75\%$ for AEGIS. The multi-link configurations have the highest observed safe-success (\cref{tab:coverage}b). On this still hazard, optical-flow transport adds no measured benefit: the multi-link configuration with a frozen hazard estimate collides in $25.19\%$ of episodes, compared to $27.00\%$ for the complete system. OSCBF collides in $66.88\%$ of static episodes (\cref{tab:coverage}b).

\begin{figure*}[t]
    
    \includegraphics[width=\linewidth]{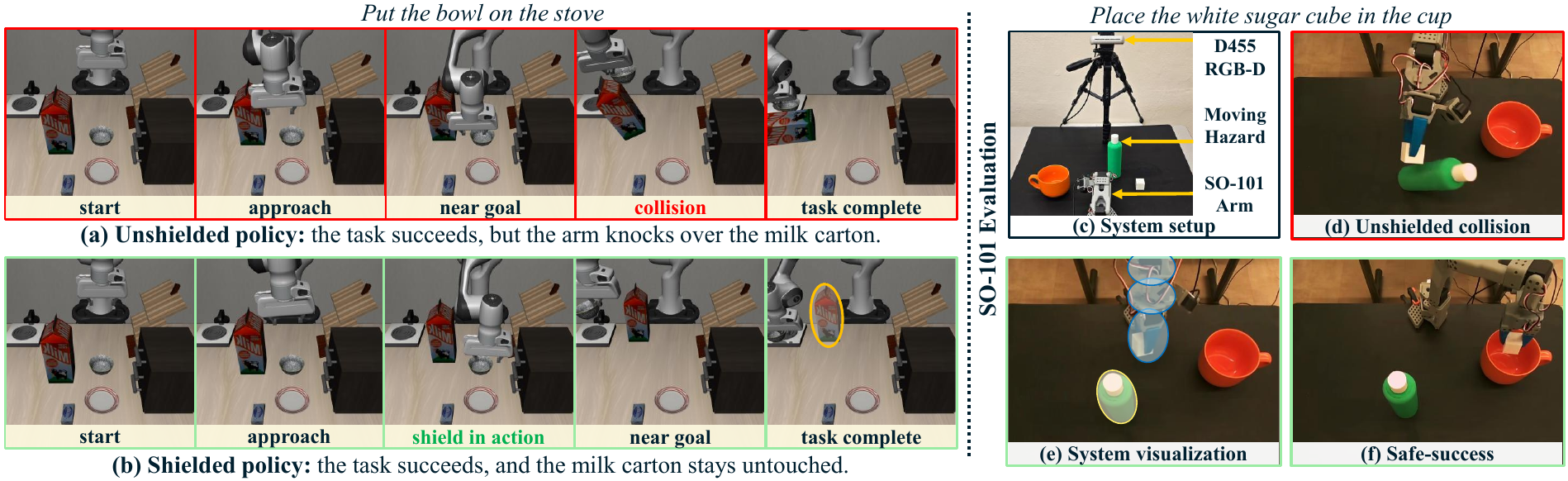}

    \vspace{-0.5em}
      
    \caption{\textbf{The shield in simulation and on hardware.} (a,b) MuJoCo, $300$ mm escape condition: both executions complete the task, but only the shielded execution leaves the moving hazard untouched, a safe-success. (c--f) One of the four tasks on a physical SO-101 arm driven by the same edge platform.}\label{fig:real}
  
    \vspace{-1.5em}
    
\end{figure*}

\subsection{Edge Deployment}
\label{sec:edge}

The barrier and the optical-flow tracker run on the CPU at their respective step cadences; the policy, the detector, and the hazard namer run on the iGPU, with the latter two running only at reset (\cref{tab:edge}). A barrier step costs $0.0189$ J, and an optical-flow update costs $0.132$ J, compared to $7.86$ J for a policy call. The detector and the hazard namer add $19.7$ J and $51.1$ J, respectively, once per episode.

In its released serving configuration, the policy takes $343$ ms per call on the iGPU. Trimming the prefix from $968$ to $576$ positions and halving the integration steps gives the deployed $177.3$ ms, with the policy weights unchanged.

On the NPU, the prefix dominates the call: prefix cache $54.3\%$ and prefix encoding $30.3\%$, compared to $15.4\%$ for the integration loop. The unbatched policy's matrix shapes suit the NPU poorly, so its call is $2.52\times$ that of the iGPU. The iGPU draws $44.0$ W against $23.8$ W on the NPU but completes each call sooner, so it spends $7.86$ J per call against $8.84$ J. The detector's fusion encoder is slower still on the NPU, at $1{,}837$ ms against $105$ ms on the iGPU.

Amortized over a ten-action chunk, with a barrier step at every control step and an optical-flow update at every fifth, the deployed mapping demands $20.3$ ms per $50$ ms control period, but each chunk boundary stalls for the duration of the policy call, $177.3$ ms at p50, when calls are issued synchronously.

\textbf{Safety under policy optimization}
For static hazards, the reduced configuration of \cref{ssec:edgelevers} leaves task success and safe-success equivalent. At the same time, collision rises by $1.75$ percentage points. With the filter active, the same reduction lowers the single-client call time by $32.7\%$ at p50.
On the $300$ mm escape condition, the same reduction leaves shielded collision at $22.92\%$ against $22.44\%$, so the shield's collision reduction is unaffected by the policy configuration. Task success falls by $4.17$ percentage points and safe-success by $3.53$ percentage points, both significant. Without the shield, the same reduction changes neither rate significantly.
The barrier corrects about a quarter of control steps in every configuration. That fraction is insensitive to the deviation threshold used to define a correction, so barrier activity is governed by geometry rather than by the policy configuration.

\textbf{Physical deployment}
We also deploy the shield on an SO-101 arm across four tabletop placement tasks, with instructions such as \textit{place the white sugar cube in the cup}, with a bottle or a box obstructing the arm's path. A fixed $\pi_{0.5}$ policy provides the nominal actions. An external Intel RealSense D455 supplies aligned RGB-D observations for fitting the hazard ellipsoid at reset and the image stream on which optical flow transports it; its RGB stream and a wrist-mounted camera provide the policy observations. 
On the same edge platform and device mapping as in \cref{tab:edge}, the transported ellipsoid follows the hazard as it is carried across the arm's path (\cref{fig:real}). Across the four tasks, four episodes each, the arm touched the hazard in $3$ of $16$ shielded episodes and in $11$ of $16$ unshielded ones; it completed the placement in $11$ shielded episodes and in $13$ unshielded ones. The hardware episode budget is small, and the hazard is hand-carried, so these counts are reported without the paired tests of \cref{sec:performance}.

\begin{table}[t]
  \centering
  \caption{\textbf{Where perception and tracking lose protection.} $300$ mm escape condition. Bold: best rate per column; $\dagger$: detector prompted with the oracle-corrected caption.}
  \label{tab:ladder}
  \footnotesize
  \setlength{\tabcolsep}{2.5pt}
  \vspace{-0.5em}

  \begin{tabular}{lrrr}
    \toprule
    Configuration & \colhead{Coll.\ (\%)$\downarrow$} & \colhead{Succ.\ (\%)$\uparrow$} & \colhead{Safe-succ.\ (\%)$\uparrow$} \\
    \midrule
    Unshielded $\pi_{0.5}$ & 45.99 & 67.63 & 42.15 \\
    Frozen hazard estimate & 30.29 & 58.81 & 45.51 \\
    Fixed-cadence refresh & 30.13 & 58.97 & 45.83 \\
    \midrule
    Exact geometry + true motion$^{\ddagger}$ & \textbf{18.43} & 65.87 & 60.10 \\
    Detector + true motion$^\dagger$ & 20.19 & \textbf{68.59} & \textbf{60.90} \\
    Detector + flow$^\dagger$ & 22.44 & 64.74 & 56.57 \\
    Ours (VLM-generated name) & 22.44 & 64.58 & 56.41 \\
    \bottomrule
  \end{tabular}
  \par\smallskip
  {\scriptsize $^{\ddagger}$Not deployable. The frozen-estimate and fixed-cadence variants use the original caption, so comparisons between the upper and lower blocks are system-level rather than tracking-only.}
  \vspace{-1.5em}
\end{table}

\section{Ablations and Analysis}
\label{sec:ablations}

This section examines the protection and task costs of multi-link coverage, losses due to perception and tracking, and the placement accuracy and update frequency required to maintain protection. Each study varies one policy factor, with the perception and controller otherwise fixed, so these are the paper's controlled comparisons along the coverage and motion axes.

\subsection{Multi-Link Protection and Its Task Cost}
\label{sec:coverage}

With perception and the controller held fixed, guarding the wrist and forearm in addition to the end effector reduces static collision from $31.31\%$ to $25.19\%$ (\cref{tab:coverage}b). Neither task success ($65.62\%$ to $61.62\%$) nor safe-success ($51.56\%$ to $52.88\%$) changes significantly. Relative to unshielded execution, safe failures (episodes that avoid the hazard but do not complete the task) rise from $1.50\%$ to $21.94\%$ of episodes under multi-link coverage. With exact hazard geometry and motion, which is not deployable, the same coverage change reduces collision from $21.94\%$ under end-effector coverage to $14.44\%$.

Shielding also reduces collision severity: unshielded episodes displace the hazard by $147.7$ mm at the median, and more than nine in ten exceed $10$ mm, whereas the disturbances surviving multi-link coverage fall to $32.5$ mm. When a collision requires $10$ mm of displacement instead of $1$ mm, multi-link coverage retains $83.7\%$ of its advantage over end-effector coverage ($5.12$ versus $6.12$ percentage points), a benefit concentrated at displacements below $50$ mm.

\begin{figure}[ht]
  \centering
  \includegraphics[width=\columnwidth]{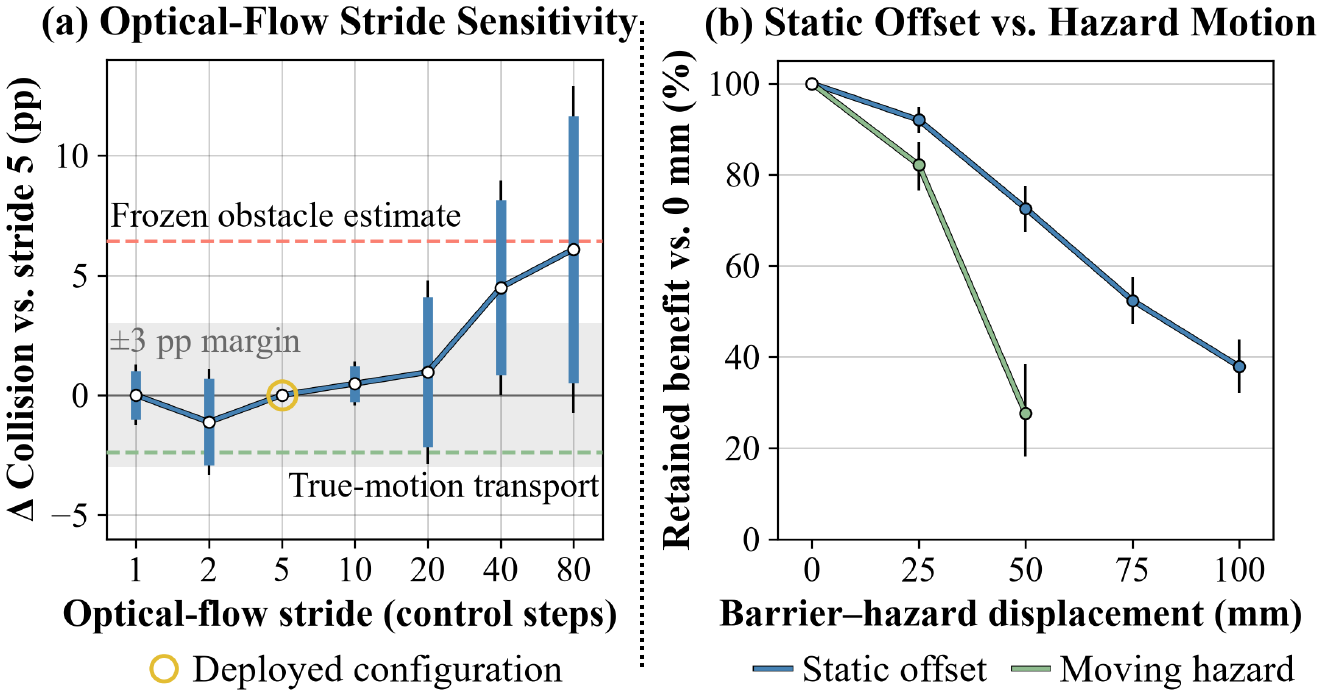}

  \vspace{-0.5em}

  \caption{(a) Per-step optical flow is equivalent to stride $5$ within the shaded $\pm3$ pp margin (pp: percentage points). (b) Motion increases placement sensitivity: the hazard leaves an ellipsoid frozen at reset; both displacement sweeps use the static benchmark. Thin bars: $95\%$ CIs; thick: $90\%$.}
  
  \label{fig:placement}
    \vspace{-1.5em}
\end{figure}

\subsection{Hazard Estimation and Tracking Failures}
\label{sec:dynamic}

Broader coverage is worth only as much as the placement of the hazard ellipsoid it works against. Exact hazard geometry and motion are therefore replaced by staged perception to locate where protection is lost. On the $300$ mm escape condition, collision falls from $30.29\%$ with a hazard estimate frozen at reset to $18.43\%$ with exact geometry and true motion, which is not deployable (\cref{tab:ladder}). Replacing the exact geometry with the detector geometry while retaining true motion raises the collision rate to $20.19\%$. At the same time, task success and safe-success are both higher, so exact geometry and motion do not upper-bound task outcomes. Replacing true motion with optical flow raises collision to $22.44\%$ and reduces task success from $68.59\%$ to $64.74\%$. Optical flow accounts for the remaining measured loss and still improves every rate over the frozen-hazard estimate. Image-based perception therefore retains most of the protection available with exact geometry and motion.

Replacing the oracle-corrected caption with the VLM-generated name leaves the aggregate collision unchanged at $22.44\%$, with negligible changes in task success and safe-success. This parity conceals a failure on the object with the weakest grounding. In six of $624$ episodes, the same plausible but incorrect phrase causes the filter guard to assign a different object, a systematic failure that aggregate rates hide.

Repeated detection incurs both computation and association costs. One detector call on the edge platform costs about $148\times$ the latency and $149\times$ the energy of an optical-flow update (\cref{tab:edge}). Refits are accepted only within $50$ mm of the last accepted position; this gate rejects correct proposals after larger motion and can retain a wrong anchor. The fixed-cadence variant therefore collides as often as a hazard estimate frozen at reset (\cref{tab:ladder}), accepting at least one refresh in only $184$ of $624$ episodes. On the stationary condition, repeated refitting raises collision by $8.81$ percentage points over the frozen estimate: the frame-to-frame shape variation that a single reset fit avoids. Optical flow instead transports the identity established at reset, updating the ellipsoid without re-detection or association.

\subsection{Placement Accuracy and Update Frequency}
\label{ssec:placement}

Two sweeps test the placement accuracy and update frequency needed to maintain protection. Retained benefit is collision reduction relative to unshielded execution, normalized by the reduction at zero displacement. Both sweeps run on the static benchmark (\cref{fig:placement}(b)).
Translating only the ellipsoid retains $92.0\%$ and $72.6\%$ of its benefit at $25$ and $50$ mm; moving the hazard away from a frozen estimate retains $82.1\%$ and $27.6\%$. The gap between the two widens from $9.9$ to $45.0$ percentage points across that interval, so a good reset fit does not replace transport during motion. The median fitted-center error on the static benchmark, $20.8$ mm, is below the smallest nonzero offset tested.

The cadence sweep uses $624$ paired $300$ mm escape episodes per configuration with the original caption, separate from the oracle-corrected-caption variants above (\cref{fig:placement}(a)). At strides $1$, $2$, $5$, $10$, and $20$, collision stays between $22.76\%$ and $24.84\%$, strides $1$ and $2$ are equivalent to stride $5$, and neither task success nor safe-success differs significantly from stride $5$. True-motion transport stays lower at $21.47\%$, and more frequent updates do not close this gap, because optical flow localizes the hazard with $48.2$ to $49.7$ mm of error at every stride, compared to $37.3$ mm with true motion. Longer strides lose protection: collision rises to $28.37\%$ at stride $40$ and $29.97\%$ at stride $80$, close to the frozen hazard estimate at $30.29\%$.

On the matched stationary condition, no stride reduces collision below the frozen estimate, and per-step measurement is equivalent to stride $5$: more frequent measurement cannot help a static hazard. Stride $5$ is therefore deployed, at $3.1$ ms per measurement on the edge CPU.

\section{Conclusion and Future Directions}\label{sec:conclusion}

This work shows that runtime protection for pretrained VLA policies can be extended beyond the end effector, kept current around moving hazards, and executed on shared edge hardware without retraining the policy. Multi-link coverage improves protection across the arm, while sparse optical flow preserves most of the benefit lost when the hazard estimate is frozen. Both the barrier and tracker remain millisecond-scale workloads, making policy inference the dominant cost of the deployed system rather than shielding.

The edge measurements also identify where that cost lies. Prefix trimming and fewer flow-matching integration steps substantially reduce policy latency, while NPU execution is still dominated by vision-language prefix encoding and cache construction. For this unbatched $\pi_{0.5}$ workload, lower accelerator power does not translate to lower energy per call; reducing prefix-processing latency is therefore the more relevant optimization target.

A remaining challenge is tighter interaction between the policy and shield. The shield modifies commanded motion, but the fixed policy does not observe those corrections within its current action chunk, which can limit recovery when avoidance requires replanning. Feeding shielded state and corrective actions back into the policy, together with safe fallback behaviors, offers a path toward higher task completion and stronger protection under moving hazards. More broadly, the results demonstrate that motion-aware, multi-link shielding can provide a practical onboard safety layer around general-purpose manipulation policies.

% The bibliography keeps IEEEtran's default spacing and full author lists.
\RemoveFromHook{para/begin}[squeeze]
\bibliographystyle{IEEEtran}
\bibliography{abbrev,references}

@String(CVPR  = {Proc. IEEE Conf. Comput. Vis. Pattern Recognit. (CVPR)})

@String(ECCV  = {Proc. Eur. Conf. Comput. Vis. (ECCV)})

@String(NeurIPS = {Adv. Neural Inf. Process. Syst. (NeurIPS)})

@String(IJCAI = {Proc. Int. Joint Conf. Artif. Intell. (IJCAI)})

@String(ICLR  = {Proc. Int. Conf. Learn. Represent. (ICLR)})

@inproceedings{pi0,
  author    = {Kevin Black and Noah Brown and Danny Driess and Adnan Esmail and Michael Equi and Chelsea Finn and Niccolo Fusai and Lachy Groom and Karol Hausman and Brian Ichter and others},
  title     = {$\pi_0$: A Vision-Language-Action Flow Model for General Robot Control},
  booktitle = {Proc. Robot.: Sci. Syst. (RSS)},
  year      = {2025},
  doi       = {10.15607/RSS.2025.XXI.010}
}

@inproceedings{pi05,
  author    = {Kevin Black and Noah Brown and James Darpinian and Karan Dhabalia and Danny Driess and Adnan Esmail and Michael Equi and Chelsea Finn and Niccolo Fusai and others},
  title     = {$\pi_{0.5}$: A Vision-Language-Action Model with Open-World Generalization},
  booktitle = {Proc. Conf. Robot Learn. (CoRL)},
  pages     = {17--40},
  year      = {2025}
}

@inproceedings{openvla,
  author    = {Moo Jin Kim and Karl Pertsch and Siddharth Karamcheti and Ted Xiao and Ashwin Balakrishna and Suraj Nair and Rafael Rafailov and Ethan Foster and Pannag Sanketi and Quan Vuong and Thomas Kollar and Benjamin Burchfiel and Russ Tedrake and Dorsa Sadigh and Sergey Levine and Percy Liang and Chelsea Finn},
  title     = {{OpenVLA}: An Open-Source Vision-Language-Action Model},
  booktitle = {Proc. Conf. Robot Learn. (CoRL)},
  pages     = {2679--2713},
  year      = {2024}
}

@inproceedings{rt2,
  author    = {Brianna Zitkovich and Tianhe Yu and Sichun Xu and Peng Xu and Ted Xiao and Fei Xia and Jialin Wu and Paul Wohlhart and others},
  title     = {{RT-2}: Vision-Language-Action Models Transfer Web Knowledge to Robotic Control},
  booktitle = {Proc. Conf. Robot Learn. (CoRL)},
  pages     = {2165--2183},
  year      = {2023}
}

@inproceedings{libero,
  author    = {Bo Liu and Yifeng Zhu and Chongkai Gao and Yihao Feng and Qiang Liu and Yuke Zhu and Peter Stone},
  title     = {{LIBERO}: Benchmarking Knowledge Transfer for Lifelong Robot Learning},
  booktitle = {Proc. Adv. Neural Inf. Process. Syst. (NeurIPS)},
  pages     = {44776--44791},
  year      = {2023},
  doi       = {10.52202/075280-1939}
}

@inproceedings{libero_safety,
  author    = {Rongxu Cui and Zongzheng Zhang and Jingrui Pang and Haohan Chi and Jinbang Guo and Saining Zhang and Shaoxuan Xie and Xin Jin and Yao Mu and Jiaolong Yang and Guocai Yao and Xianyuan Zhan and Ya-Qin Zhang and Hao Zhao},
  title     = {{LIBERO-Safety}: A Comprehensive Benchmark for Physical and Semantic Safety in Vision-Language-Action Models},
  booktitle = {Proc. Eur. Conf. Comput. Vis. (ECCV)},
  pages     = {512--530},
  year      = {2026},
  doi       = {10.1007/978-3-032-37627-5_28}
}

@inproceedings{groundingdino,
  author    = {Shilong Liu and Zhaoyang Zeng and Tianhe Ren and Feng Li and Hao Zhang and Jie Yang and Qing Jiang and Chunyuan Li and Jianwei Yang and Hang Su and Jun Zhu and Lei Zhang},
  title     = {Grounding {DINO}: Marrying {DINO} with Grounded Pre-Training for Open-Set Object Detection},
  booktitle = {Proc. Eur. Conf. Comput. Vis. (ECCV)},
  pages     = {38--55},
  year      = {2024},
  doi       = {10.1007/978-3-031-72970-6_3}
}

@inproceedings{sam2,
  author    = {Nikhila Ravi and Valentin Gabeur and Yuan-Ting Hu and Ronghang Hu and Chaitanya Ryali and Tengyu Ma and Haitham Khedr and Roman R{\"a}dle and Chloe Rolland and Laura Gustafson and Eric Mintun and Junting Pan and Kalyan Vasudev Alwala and Nicolas Carion and Chao-Yuan Wu and Ross Girshick and Piotr Doll{\'a}r and Christoph Feichtenhofer},
  title     = {{SAM} 2: Segment Anything in Images and Videos},
  booktitle = {Proc. Int. Conf. Learn. Represent. (ICLR)},
  year      = {2025}
}

@article{ames_cbf,
  author  = {Aaron D. Ames and Xiangru Xu and Jessy W. Grizzle and Paulo Tabuada},
  title   = {Control Barrier Function Based Quadratic Programs for Safety Critical Systems},
  journal = {IEEE Trans. Autom. Control},
  volume  = {62},
  number  = {8},
  pages   = {3861--3876},
  month   = aug,
  year    = {2017},
  doi     = {10.1109/TAC.2016.2638961}
}

@inproceedings{cbf_qp,
  author    = {Aaron D. Ames and Samuel Coogan and Magnus Egerstedt and Gennaro Notomista and Koushil Sreenath and Paulo Tabuada},
  title     = {Control Barrier Functions: Theory and Applications},
  booktitle = {Proc. Eur. Control Conf. (ECC)},
  pages     = {3420--3431},
  year      = {2019},
  doi       = {10.23919/ECC.2019.8796030}
}

@inproceedings{mrcbf,
  author    = {Sarah Dean and Andrew J. Taylor and Ryan K. Cosner and Benjamin Recht and Aaron D. Ames},
  title     = {Guaranteeing Safety of Learned Perception Modules via Measurement-Robust Control Barrier Functions},
  booktitle = {Proc. Conf. Robot Learn. (CoRL)},
  pages     = {654--670},
  year      = {2020}
}

@article{rotating_hyperplane_cbf,
  author  = {Riku Funada and Koju Nishimoto and Tatsuya Ibuki and Mitsuji Sampei},
  title   = {Collision Avoidance for Ellipsoidal Rigid Bodies With Control Barrier Functions Designed From Rotating Supporting Hyperplanes},
  journal = {IEEE Trans. Control Syst. Technol.},
  volume  = {33},
  number  = {1},
  pages   = {148--164},
  month   = jan,
  year    = {2025},
  doi     = {10.1109/TCST.2024.3467809}
}

@article{sampled_wholebody_cbf,
  author  = {Yuhan Xiong and Di-Hua Zhai and Yuanqing Xia},
  title   = {Robust Whole-Body Safety-Critical Control for Sampled-Data Robotic Manipulators via Control Barrier Functions},
  journal = {IEEE Trans. Autom. Sci. Eng.},
  volume  = {22},
  pages   = {16050--16061},
  year    = {2025},
  doi     = {10.1109/TASE.2025.3574342}
}

@misc{vlsa_aegis,
  author  = {Songqiao Hu and Zeyi Liu and Shuang Liu and Jun Cen and Zihan Meng and Shihefeng Wang and Xiang Li and Xiao He},
  title   = {{VLSA}: Vision-Language-Action Models with Plug-and-Play Safety Constraint Layer},
  year    = {2025},
  note    = {{arXiv}:2512.11891},
  doi     = {10.48550/arXiv.2512.11891}
}

@inproceedings{oscbf,
  author    = {Daniel Morton and Marco Pavone},
  title     = {Safe, Task-Consistent Manipulation with Operational Space Control Barrier Functions},
  booktitle = {Proc. IEEE/RSJ Int. Conf. Intell. Robots Syst. (IROS)},
  pages     = {187--194},
  year      = {2025},
  doi       = {10.1109/IROS60139.2025.11246389}
}

@misc{agsf_knows,
  author  = {Seongbin Park and Fan Zhang and Baharan Mirzasoleiman and Shahriar Talebi and Nader Sehatbakhsh},
  title   = {Your Model Already Knows: Attention-Guided Safety Filter for Vision-Language-Action Models},
  year    = {2026},
  note    = {{arXiv}:2606.09749},
  doi     = {10.48550/arXiv.2606.09749}
}

@misc{nsym_flow,
  author  = {William English and Hao Zheng and Rickard Ewetz},
  title   = {Neuro-Symbolic Safety Guidance for Vision-Language-Action Models via Constrained Flow Matching},
  year    = {2026},
  note    = {{arXiv}:2607.01378},
  doi     = {10.48550/arXiv.2607.01378}
}

@misc{barrier_flow,
  author  = {Kasra Sinaei and Hung-Chieh Wu and Donald Ebeigbe},
  title   = {Safe Vision Language Action Models via Barrier Enhanced Flow Matching},
  year    = {2026},
  note    = {{arXiv}:2607.29569},
  doi     = {10.48550/arXiv.2607.29569}
}

@misc{anybody_guard,
  author  = {Alex Beaudin and Hanna Krasowski and Kartik Nagpal and Sanjit A. Seshia and Murat Arcak and Negar Mehr},
  title   = {{Any-Body Guard}: Universal Safeguarding for Manipulation Policies via Action Masking},
  year    = {2026},
  note    = {{arXiv}:2606.22278},
  doi     = {10.48550/arXiv.2606.22278}
}

@misc{smolvla,
  author  = {Mustafa Shukor and Dana Aubakirova and Francesco Capuano and Pepijn Kooijmans and Steven Palma and Adil Zouitine and Michel Aractingi and Caroline Pascal and Martino Russi and Andres Marafioti and others},
  title   = {{SmolVLA}: A Vision-Language-Action Model for Affordable and Efficient Robotics},
  year    = {2025},
  note    = {{arXiv}:2506.01844},
  doi     = {10.48550/arXiv.2506.01844}
}

@misc{edgevla,
  author  = {Pawe{\l} Budzianowski and Wesley Maa and Matthew Freed and Jingxiang Mo and Winston Hsiao and Aaron Xie and Tomasz M{\l}oduchowski and Viraj Tipnis and Benjamin Bolte},
  title   = {{EdgeVLA}: Efficient Vision-Language-Action Models},
  year    = {2025},
  note    = {{arXiv}:2507.14049},
  doi     = {10.48550/arXiv.2507.14049}
}

@inproceedings{openvla_oft,
  author    = {Moo Jin Kim and Chelsea Finn and Percy Liang},
  title     = {Fine-Tuning Vision-Language-Action Models: Optimizing Speed and Success},
  booktitle = {Proc. Robot.: Sci. Syst. (RSS)},
  year      = {2025},
  doi       = {10.15607/RSS.2025.XXI.017}
}

@inproceedings{quantvla,
  author    = {Jingxuan Zhang and Yunta Hsieh and Zhongwei Wan and Haokun Lin and Xin Wang and Ziqi Wang and Yingtie Lei and Mi Zhang},
  title     = {{QuantVLA}: Scale-Calibrated Post-Training Quantization for Vision-Language-Action Models},
  booktitle = {Proc. IEEE/CVF Conf. Comput. Vis. Pattern Recognit. (CVPR)},
  year      = {2026}
}

@article{bfapp,
  author  = {Haosheng Li and Weixin Mao and Zihan Lan and Hongwei Xiong and Hongan Wang and Chenyang Si and Ziwei Liu and Xiaoming Deng and Hua Chen},
  title   = {{BFA++}: Hierarchical Best-Feature-Aware Token Prune for Multi-View Vision Language Action Model},
  journal = {IEEE Robot. Autom. Lett.},
  volume  = {11},
  number  = {5},
  pages   = {6002--6009},
  month   = may,
  year    = {2026},
  doi     = {10.1109/LRA.2026.3673974}
}

@misc{xpu_char,
  author  = {Kaijun Zhou and Qiwei Chen and Da Peng and Zhiyang Li and Xijun Li and Jinyu Gu},
  title   = {Characterizing Vision-Language-Action Models across {XPUs}: Constraints and Acceleration for On-Robot Deployment},
  year    = {2026},
  note    = {{arXiv}:2604.24447},
  doi     = {10.48550/arXiv.2604.24447}
}

@inproceedings{lucaskanade,
  author    = {Bruce D. Lucas and Takeo Kanade},
  title     = {An Iterative Image Registration Technique with an Application
               to Stereo Vision},
  booktitle = {Proc. Int. Joint Conf. Artif. Intell. (IJCAI)},
  pages     = {674--679},
  year      = {1981}
}

@article{holm,
  author  = {Sture Holm},
  title   = {A Simple Sequentially Rejective Multiple Test Procedure},
  journal = {Scand. J. Statist.},
  volume  = {6},
  number  = {2},
  pages   = {65--70},
  year    = {1979}
}

\end{document}